\documentclass[sigconf]{acmart}
\usepackage{multirow} 

\usepackage{algpseudocode}
\usepackage{algorithm}
\usepackage{multicol}
\usepackage{array}
\usepackage{arydshln}
\usepackage{booktabs}  
\usepackage{float}
\usepackage{color}

\begin{document}

\title{Docking-Aware Attention: Dynamic Protein Representations through Molecular Context Integration}

\author{Amitay Sicherman}
\email{amitay.s@cs.technion.ac.il}
\affiliation{%
  \institution{Technion - Israel Institute of Technology}
  \country{Israel}
}

\author{Kira Radinsky}
\email{kirar@cs.technion.ac.il}
\affiliation{
  \institution{Technion - Israel Institute of Technology}
  \country{Israel}
}

\begin{abstract}

Computational prediction of enzymatic reactions represents a crucial challenge in sustainable chemical synthesis across various scientific domains, ranging from drug discovery to materials science and green chemistry. These syntheses rely on proteins that selectively catalyze complex molecular transformations. These protein catalysts exhibit remarkable substrate adaptability, with the same protein often catalyzing different chemical transformations depending on its molecular partners. Current approaches to protein representation in reaction prediction either ignore protein structure entirely or rely on static embeddings, failing to capture how proteins dynamically adapt their behavior to different substrates. We present Docking-Aware Attention (DAA), a novel architecture that generates dynamic, context-dependent protein representations by incorporating molecular docking information into the attention mechanism. DAA combines physical interaction scores from docking predictions with learned attention patterns to focus on protein regions most relevant to specific molecular interactions. We evaluate our method on enzymatic reaction prediction, where it outperforms previous state-of-the-art methods, achieving  62.2\% accuracy versus 56.79\% on complex molecules and 55.54\% versus 49.45\% on innovative reactions. Through detailed ablation studies and visualizations, we demonstrate how DAA generates interpretable attention patterns that adapt to different molecular contexts. Our approach represents a general framework for context-aware protein representation in biocatalysis prediction, with potential applications across enzymatic synthesis planning. We open-source our implementation and pre-trained models to facilitate further research.

\end{abstract}


\maketitle

\section{Introduction}

Developing effective computational approaches for enzymatic reaction prediction is a fundamental challenge in machine learning for chemistry \cite{goshisht2024machine,kreutter2021predicting}. The ability to accurately predict enzyme-catalyzed reactions has significant implications across multiple scientific domains, from drug discovery and metabolic engineering to sustainable chemistry and materials science \cite{sen2015green,fessner2015systems}. Enzymes, as protein-based catalysts, enable complex molecular transformations under mild conditions with high selectivity and efficiency. However, the same enzyme can exhibit different catalytic behaviors depending on its molecular partners, making computational modeling of these reactions particularly challenging.

This substrate-dependent adaptability of enzymes poses unique challenges for machine learning models. Current computational methods for enzymatic reaction prediction either ignore protein structure entirely or rely on enzyme commission (EC) numbers for protein representation~\cite{Probst2022,Chen2023}. While EC numbers provide useful categorical information about enzyme function, they fail to capture the nuanced ways proteins interact with different molecular partners. Even advanced deep learning approaches, whether using protein language models like ESM \cite{hayes2024simulating} or structure-based models like GearNet \cite{zhang2023protein}, face a fundamental limitation—while they effectively capture sequence or structural properties, they still produce static protein representations that fail to reflect the protein's dynamic behavior with different molecules in their environment.

Molecular docking has long been used to study protein-ligand interactions \cite{muhammed2024molecular}, providing physics-based insights into binding preferences and interaction patterns. Recent advances in deep learning have improved docking accuracy and efficiency \cite{DiffDock2023}, making it feasible to incorporate this valuable source of structural interaction information into protein representation learning. However, previous attempts to combine docking with deep learning have primarily focused on improving binding pose prediction rather than generating dynamic protein representations for reaction prediction.

We present Docking-Aware Attention (DAA), a novel architecture that addresses these limitations by generating dynamic, context-dependent protein representations. Our key insight is that molecular docking information can guide attention mechanisms to focus on protein regions most relevant to specific molecular interactions. By incorporating docking scores into the attention computation, DAA produces protein representations that adapt based on the predicted physical interactions between the protein and its molecular partners. This approach better reflects the reality of enzyme behavior, where catalytic activity depends on specific substrate interactions.

The key contributions of this work are: (i) We introduce DAA, a novel architecture that integrates physical protein-ligand interaction information into attention mechanisms to generate context-dependent protein representations for enzymatic reaction prediction; (ii) Through extensive experiments on biocatalysis prediction, we show that these context-aware protein representations substantially improve performance on challenging cases, achieving 62.2\% versus 56.79\% accuracy on complex molecules and 55.54\% versus 49.45\% on innovative reactions; (iii) We demonstrate how DAA generates interpretable attention patterns that reveal which protein regions are most relevant for specific molecular interactions, providing insights into context-dependent enzyme behavior.

To facilitate reproducibility and encourage further development in this direction, we have open-sourced our complete codebase, including model implementations, training scripts, and pre-trained models at GitHub \footnote{\url{https://anonymous.4open.science/r/DockingAwareAttention-8B8E}}.

\section{Related Works}

\subsection{Biocatalysis Prediction}
The evolution of biocatalysis prediction has been significantly shaped by advances in chemical reaction prediction, particularly in sequence-to-sequence models. Neural sequence-to-sequence approaches \cite{schwaller2019molecular} pioneered the treatment of reactions as translation tasks, with transformer architectures like ChemBERTa \cite{ChithranandaChemBERTa} and MolFormer \cite{ross2022large} further improving the capture of molecular dependencies.

The adaptation of these models to biocatalysis presented unique challenges in representing enzyme-substrate interactions. Kreutter et al. \cite{kreutter2021predicting} first demonstrated the viability of transformer models for biocatalysis by leveraging upsampled enzymatic reaction data. ECREACT \cite{Probst2022} advanced this approach by representing enzymes through EC numbers as special tokens in the sequence-to-sequence framework. However, this discrete representation limited the model's ability to capture nuanced enzyme-substrate interactions.

Our work addresses these limitations through the DAA mechanism, which creates dynamic, substrate-specific enzyme representations within the sequence-to-sequence framework, enabling more precise modeling of enzyme-substrate interactions.

\subsection{Representation Learning in Proteins}
Protein representation learning has developed along sequence-based and structure-based approaches.

\subsubsection{Sequence-based Protein Language Models}
Significant advances in protein language models have been seen in recent years, with transformer-based architectures\cite{Vaswani2017} leading the way. Notable developments include ProteinBERT \cite{brandes2021proteinbert} and ProtTrans\cite{elnaggar2021prottrans}, which adapted BERT\cite{devlin2018bert} for protein sequences, and the ESM model family \cite{hayes2024simulating, lin2023evolutionary, rives2021biological}, which demonstrated the benefits of scale in protein modeling. These models have shown remarkable success in capturing local and long-range protein interactions, establishing new benchmarks in protein property prediction tasks.

\subsubsection{Structure-based Representation Learning}
Following AlphaFold's \cite{jumper2021highly} breakthrough in protein structure prediction, structure-based representation learning has gained prominence. Models like GearNet \cite{zhang2023protein} have introduced novel architectures for capturing protein structure through graph-based approaches. Various methods have emerged to better model protein geometry and structural relationships \cite{fan2023continuous, hermosilla2020intrinsic, jing2020learning}, highlighting the importance of three-dimensional information in protein understanding.

Despite these advances, both sequence-based and structure-based approaches generate static protein representations that remain fixed regardless of molecular context, limiting their ability to capture the dynamic nature of protein-molecule interactions.

\subsection{Molecular Docking Approaches}
The evolution of molecular docking methods provides crucial context for our work. Traditional approaches relied on physics-based scoring functions and search algorithms \cite{McNutt2021,Stark2022}, facing computational challenges particularly in blind docking scenarios. Recent learning-based methods have made significant strides in addressing these limitations. EquiBind \cite{stark2022equibind} introduced keypoint-based methods for pocket-ligand alignment, while TANKBind \cite{lu2022tankbind} enabled independent predictions for multiple binding sites. DiffDock \cite{DiffDock2023} represented a paradigm shift by reformulating docking as a generative modeling problem. While these approaches have improved docking accuracy and efficiency, they primarily focus on predicting binding poses rather than using docking information to enhance protein representations. Our DAA method uniquely leverages docking predictions to create dynamic protein embeddings that adapt to different molecular interaction contexts.

%%%%%%%%%%%%%%%%%%%%%%%%%%%%%%%%%%%%%%%%%%%%%%%%%%%%%%%%%%%%%%%%
%%%%%%%%%%%%%%%%        ALGO  %%%%%%%%%%%%%%%%%%%%%%%%%%%
%%%%%%%%%%%%%%%%%%%%%%%%%%%%%%%%%%%%%%%%%%%%%%%%%%%%%%%%%%%%%%%%
\begin{algorithm}[ht]
\caption{Docking-Aware Attention (DAA)}
\label{algo:daa}
\begin{algorithmic}[1]
\Require Protein  $P$, molecule $M$
\Ensure Context-dependent protein representation $\mathbf{p}_M$
\State // Convert protein to embeddings
\State $\mathbf{E} = [\mathbf{e}_1, \ldots, \mathbf{e}_n] \gets \text{ESM}(P)$

\State // Get interaction scores from multiple docking poses
\For
   \State $\{\mathbf{p}_i^{3D}, \mathbf{m}_j^{3D}\}_k \gets \text{DiffDock}(P, M)$
   \State $V_i^k \gets \sum_j \text{LJ-potential}(\mathbf{p}_i^{3D}, \mathbf{m}_j^{3D})$
\EndFor
\State $V_i \gets \frac{1}{K}\sum_k V_i^k$ 

\State // Smooth interaction scores
\State $\hat{V}_i \gets \beta V_i + (1-\beta)\frac{1}{n}\sum_j V_j$

\State // Compute context-aware representation
\State $\mathbf{p}_M \gets \text{softmax}\left(\frac{\mathbf{Q}\mathbf{K}^T + \gamma\hat{\mathbf{V}}}{\sqrt{d}}\right)\mathbf{V}$

\State \textbf{return} $\mathbf{p}_M$
\end{algorithmic}
\end{algorithm}

%%%%%%%%%%%%%%%%%%%%%%%%%%%%%%%%%%%%%%%%%%%%%%%%%%%%%%%%%%%%%%%%
%%%%%%%%%%%%%%%%  ALGO  %%%%%%%%%%%%%%%%%%%%%%%%%%%
%%%%%%%%%%%%%%%%%%%%%%%%%%%%%%%%%%%%%%%%%%%%%%%%%%%%%%%%%%%%%%%%
%%%%%%%%%%%%%%%%%%%%%%%%%%%%%%%%%%%%%%%%%%%%%%%%%%%%%%%%%%%%%%%%
%%%%%%%%%%%%%%%%        Main Figure  %%%%%%%%%%%%%%%%%%%%%%%%%%%
%%%%%%%%%%%%%%%%%%%%%%%%%%%%%%%%%%%%%%%%%%%%%%%%%%%%%%%%%%%%%%%%
\begin{figure*}[t]
    \includegraphics[width=0.85\textwidth]{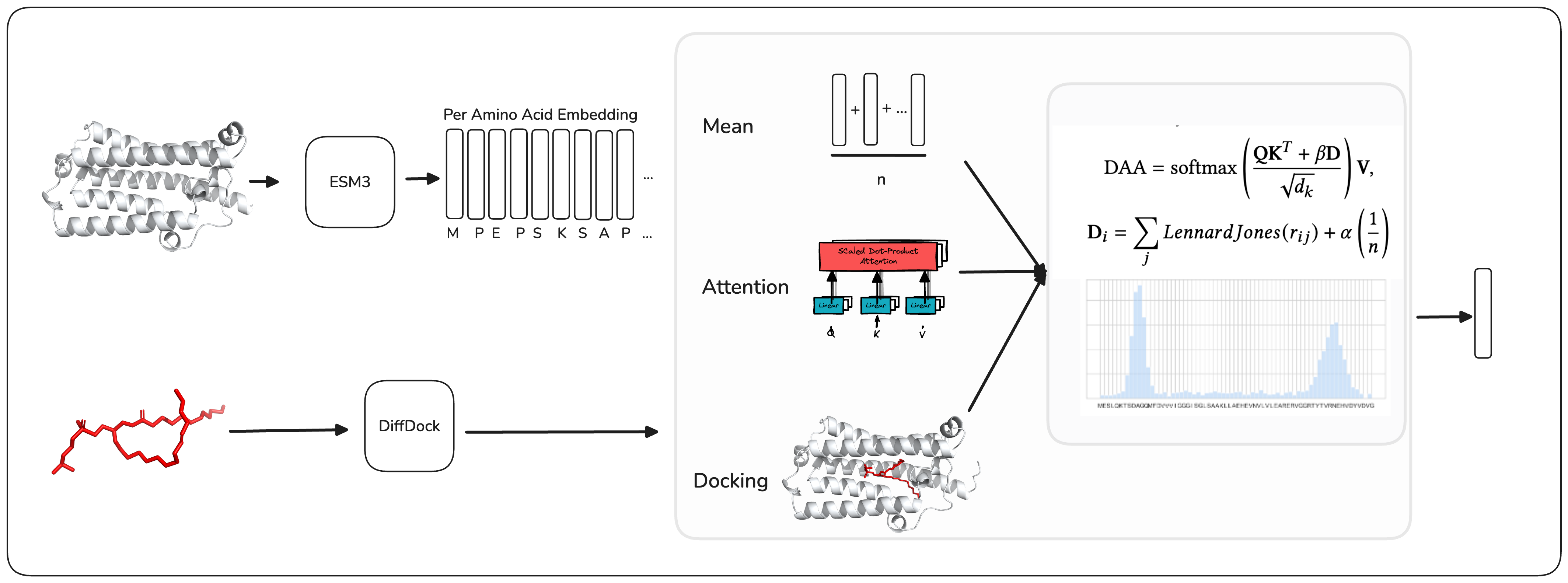}
\caption{Overview of the DAA architecture. The protein sequence is processed through a pre-trained language model for per-amino acid embeddings. The DAA mechanism integrates sequence pooling, docking scores, and learned weights to create context-aware attention, which produces a final protein representation that incorporates protein-ligand interaction information}    \label{fig:overview}
\end{figure*}
%%%%%%%%%%%%%%%%%%%%%%%%%%%%%%%%%%%%%%%%%%%%%%%%%%%%%%%%%%%%%%%%
%%%%%%%%%%%%%%%%  END Main Figure  %%%%%%%%%%%%%%%%%%%%%%%%%%%
%%%%%%%%%%%%%%%%%%%%%%%%%%%%%%%%%%%%%%%%%%%%%%%%%%%%%%%%%%%%%%%%
\section{Docking-Aware Attention}
Proteins are highly dynamic molecules whose function and behavior change based on their molecular environment. For example, enzymes can catalyze different reactions depending on the substrate molecules present, and transcription factors can change their binding behavior based on the presence of different cofactors. Despite this inherent contextual nature, current protein representation learning approaches - based on sequence or structure - generate static embeddings that remain fixed regardless of the molecular context. This fundamental limitation means that the same protein receives identical representation, whether it is interacting with different substrates, cofactors, or regulatory molecules.

We present Docking-Aware Attention (DAA), a novel architecture that addresses this limitation by generating dynamic, context-dependent protein representations. Our key insight is that protein function is intimately tied to its molecular interactions, and these interactions can be predicted through molecular docking. By incorporating docking information into the attention mechanism, DAA generates protein representations that adapt based on the specific molecules in the protein's environment. This approach better reflects the reality of protein behavior, where the same protein can exhibit different properties and functions depending on its molecular context.

\subsection{Methodological Framework}
Figure \ref{fig:overview} provides a detailed architectural overview of our approach, which is formally presented in Algorithm \ref{algo:daa}. The algorithm takes as input a protein $P$ and a molecule $M$, generating a context-dependent protein representation $\mathbf{p}_M$ through the following steps:

First, we leverage the ESM protein language model to encode the protein sequence (line 2). This encoding generates per-residue embeddings $\mathbf{E}$, providing a rich foundation of protein sequence information that will be further refined through our attention mechanism (Section \ref{sec:base_encoder}).

The molecular interaction phase (lines 4-7) employs DiffDock's generative capabilities to sample $K$ distinct binding poses between the protein and molecule. For each sampled configuration, we compute interaction scores using the Lennard-Jones potential, capturing the physical basis of protein-molecule interactions. These scores are then averaged across all poses to obtain robust estimates of interaction strength (Section \ref{sec:mol_interaction}).

The interaction scores undergo adaptive smoothing (lines 8-10) through a weighted combination of local and global information, ensuring a balance between position-specific signals and overall protein context. This smoothed interaction profile guides our attention mechanism in focusing on regions most relevant to the specific molecular interaction (Section \ref{sec:mol_smooth}).

Finally, our novel attention mechanism (line 12) integrates the smoothed interaction scores with the protein embeddings to generate a context-dependent representation $\mathbf{p}_M$. This mechanism, detailed in Section \ref{sec:daa_mechanism}, learns to balance physical interaction information with sequence-based patterns, producing protein representations that dynamically adapt to different molecular contexts.

\subsection{Architectural Components}

\subsubsection{Base Protein Encoder}
\label{sec:base_encoder}
We use ESM3-6B as our base protein encoder, leveraging its state-of-the-art performance in protein representation learning. Given a protein sequence $P$, the encoder generates embeddings for each amino acid:

\begin{equation}
    \mathbf{E} = [\mathbf{e}_1, ..., \mathbf{e}_n] = \text{ESM3}(P)
\end{equation}

where $\mathbf{e}_i \in \mathbb{R}^d$ represents the embedding of the $i$-th amino acid, and $n$ is the sequence length.

\subsubsection{Molecular Interaction Module}
\label{sec:mol_interaction}
In biological systems, proteins and molecules exist as three-dimensional structures that physically interact in space. Molecular docking predicts how a molecule will position itself and bind to a protein - essentially determining the 3D coordinates where the molecule and protein will form a complex. This spatial arrangement is crucial as it determines the strength and nature of their interaction.

For predicting these protein-molecule interactions, we employ DiffDock \cite{DiffDock2023}, a state-of-the-art molecular docking method that achieves superior accuracy compared to traditional approaches. A key advantage of DiffDock is its foundation in diffusion-based generative modeling, which allows us to sample multiple possible binding configurations for each protein-molecule pair. Unlike traditional docking methods that produce a single prediction, DiffDock generates a distribution of potential binding poses:

\begin{equation}
    \{\mathbf{p}_i^{3D}, \mathbf{m}_j^{3D}\}_k = \text{DiffDock}(P, M, k), \quad k = 1,...,K
\end{equation}

Where $k$ indexes the different sampled configurations and $K$ is the number of samples. This sampling approach better reflects the dynamic nature of protein-molecule interactions and provides a more robust estimation of binding preferences.

Once we have multiple protein-molecule configurations in the same coordinate space, our next goal is to determine which amino acids are most important for this specific protein-molecule interaction. The key insight is that amino acids that interact strongly with the molecule across multiple binding poses are likely to play crucial roles in the protein's function regarding that specific molecule.

To quantify these interaction strengths, we use the Lennard-Jones potential~\cite{tee1966molecular,wang2020lennard}, a well-established approximation in molecular physics that captures both attractive and repulsive forces between particles. The potential includes a repulsive term ($r^{-12}$) that dominates at short interatomic distances, modeling strong short-range repulsion due to quantum mechanical exchange interactions and the Pauli exclusion principle. Additionally, it includes an attractive term ($r^{-6}$) that represents van der Waals interactions, which arise from induced dipole-dipole (London dispersion) forces. While the Lennard-Jones potential is an empirical approximation, it effectively captures these essential features of intermolecular interactions. For each sampled molecular configuration, we compute the system's total potential energy as the sum of all pairwise Lennard-Jones interactions:

\begin{equation}
    S_i = \frac{1}{K}\sum_{k=1}^K \sum_{j=1}^{n_m} 4\varepsilon \left[\left(\frac{\sigma}{r_{ij}^k}\right)^{12} - \left(\frac{\sigma}{r_{ij}^k}\right)^{6}\right]
\end{equation}

where:
\begin{itemize}
    \item $r_{ij}^k = \|\mathbf{p}_i^{3D} - \mathbf{m}_j^{3D}\|_2$ is the distance between amino acid $i$ and molecule atom $j$ in the $k$-th sampled configuration
    \item $\varepsilon$ determines the depth of the potential well 
    \item $\sigma$ is the distance at which the potential becomes zero
\end{itemize}

The resulting score $S_i$ provides a physics-based measure of the interaction strength between each amino acid and the molecule averaged across multiple possible binding modes. Higher absolute values of $S_i$ indicate stronger consistent interactions across different binding poses, suggesting that amino acid $i$ plays a more important role in this specific protein-molecule interaction.

These ensemble-averaged interaction scores will later bias our attention mechanism, allowing the model to focus more on amino acids that are physically relevant to the specific molecular interaction being considered.

\subsubsection{Adaptive Score Smoothing}
\label{sec:mol_smooth}
To balance local and global information, we introduce an adaptive smoothing mechanism:

\begin{equation}
    \hat{S}_i = \beta V_i + (1-\beta) \cdot \frac{1}{n}\sum_{j=1}^n S_j
\end{equation}

The learnable parameter $\beta$ allows the model to automatically determine the optimal balance between local interaction signals and global protein context.

\subsection{Docking-Aware Attention Mechanism}
\label{sec:daa_mechanism}
Finally, we use the smoothed interaction scores to guide an attention mechanism combining amino acid embeddings into a context-dependent protein representation. Our modified attention mechanism is defined as:

\begin{equation}
    \text{Attention}(\mathbf{Q}, \mathbf{K}, \mathbf{V}, \mathbf{S}) = \text{softmax}\left(\frac{\mathbf{Q}\mathbf{K}^T + \gamma{\mathbf{S}}}{\sqrt{d}}\right)\mathbf{V}
\end{equation}

where $\mathbf{Q}, \mathbf{K}, \mathbf{V}$ are query, key, and value matrices that the model learns during training, ${\mathbf{S}}$ contains the smoothed interaction scores which vary based on the specific molecule being considered, and $\gamma$ is a parameter that learns how to balance docking information with learned attention patterns.

Unlike standard attention mechanisms that only learn from sequence patterns, our model learns during training how to effectively combine learned attention ($\mathbf{Q}\mathbf{K}^T$) with physical interaction information ($\mathbf{S}$). The trained parameters ($\mathbf{Q}, \mathbf{K}, \mathbf{V}$, and $\gamma$) become fixed, learning optimal patterns for incorporating docking information. During inference, while these parameters remain constant, the docking scores $\mathbf{S}$ vary based on each new molecule. This allows our attention mechanism to generate molecule-specific attention patterns by combining fixed learned weights with dynamic physical interaction scores - essentially creating protein representations that adapt to each molecular partner while preserving the general patterns learned during training.

%%%%%%%%%%%%%%%%%%%%%%%%%%%%%%%%%%%%%%%%%%%%%%%%%%%%%%%%%%%%%%%%
%%%%%%%%%%%%%%%%        ARC Figure  %%%%%%%%%%%%%%%%%%%%%%%%%%%
%%%%%%%%%%%%%%%%%%%%%%%%%%%%%%%%%%%%%%%%%%%%%%%%%%%%%%%%%%%%%%%%
\begin{figure*}[ht]
    \includegraphics[width=0.95\textwidth]{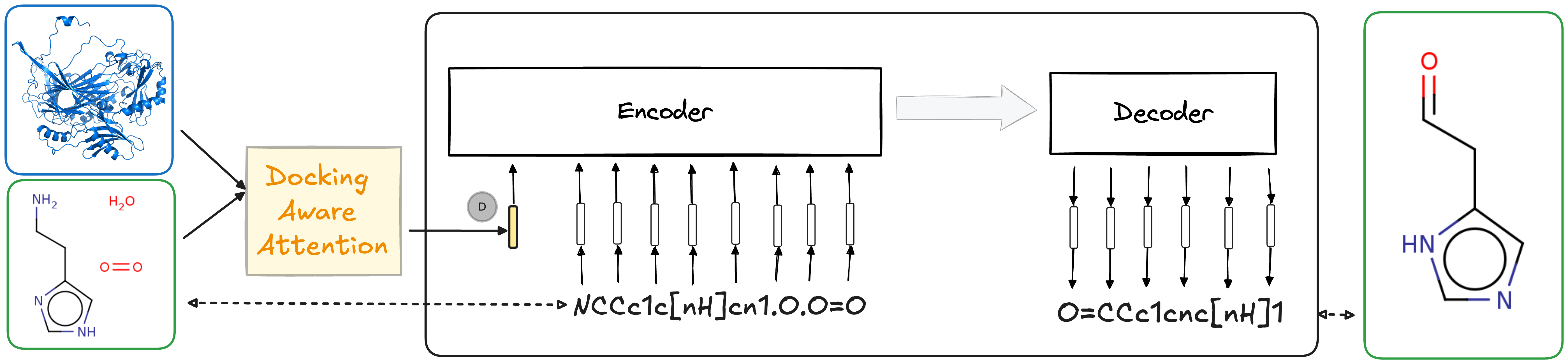}
    \caption{Overview of the biocatalysis generation pipeline. The model takes as input a catalyst enzyme and input molecule in SMILES format. These inputs are processed through our Docking-Aware Attention (DAA) mechanism to generate a molecule-specific protein representation. This representation is incorporated as a special token in the encoder,  which processes the input SMILES string. The decoder then predicts the output molecule's SMILES string, representing the reaction product.}
    \label{fig:biocat}
\end{figure*}

%%%%%%%%%%%%%%%%%%%%%%%%%%%%%%%%%%%%%%%%%%%%%%%%%%%%%%%%%%%%%%%%
%%%%%%%%%%%%%%%%      end  Figure  %%%%%%%%%%%%%%%%%%%%%%%%%%%
%%%%%%%%%%%%%%%%%%%%%%%%%%%%%%%%%%%%%%%%%%%%%%%%%%%%%%%%%%%%%%%%
\section{Empirical Evaluation}
\subsection{Task}
Biocatalysis prediction represents a fundamental task in computational biology, where enzymes catalyze chemical reactions under milder conditions with higher specificity compared to traditional chemical catalysis. The core task involves predicting the product molecule given an input molecule and its catalyzing enzyme.

We utilize ECREACT \cite{Probst2022}, currently the largest publicly available biocatalysis dataset, which aggregates enzymatic reactions from major biochemical databases including Rhea \cite{bansal2022rhea}, BRENDA \cite{placzek2016brenda}, PathBank \cite{wishart2020pathbank}, and MetaNetX \cite{ganter2013metanetx}. The dataset provides approximately 60,000 unique reactions, each containing input molecule, output molecule, and catalyzing enzyme information.

A significant challenge in enzymatic reaction prediction is the limited availability of training data compared to general chemical reactions. Following established practice in the field \cite{Probst2022,Chen2023,kreutter2021predicting,goshisht2024machine}, enzymatic reaction datasets are typically augmented with the USPTO dataset \cite{marco2015uspto}, which contains approximately 1 million non-enzymatic reactions. This augmentation commonly employs upsampling of non-enzymatic to enzymatic reactions. Previous works have demonstrated that this data augmentation strategy significantly improves model generalization by exposing the models to a broader range of chemical transformations, with multiple studies reporting enhanced performance across various reaction types \cite{Probst2022,Chen2023}.

\subsection{Baseline Methods}

We evaluate our approach against a diverse set of baseline methods, spanning three key categories: approaches that use no protein information, methods that rely on supervised EC number classification for discrete protein representation, and approaches that employ dense protein representations.

The first category of baselines operates without explicit protein information. The Chemical-Only baseline \cite{kreutter2021predicting} employs T5 architecture to predict reactions based solely on SMILES strings, relying on the model to learn enzyme-specific patterns purely from the reaction data. 
For methods relying on supervised enzyme classification, the EC Token method \cite{Probst2022} represents enzymes using their enzyme commission (EC) numbers as special tokens, requiring supervised enzyme classification information and limiting generalization to novel proteins that haven't been manually annotated with EC numbers.
For methods incorporating protein information through dense representations, we evaluate several approaches. ESM3 \cite{hayes2024simulating} and ProtBERT \cite{brandes2021proteinbert} utilize protein language models to generate sequence-based embeddings, while GearNet \cite{zhang2023protein} incorporates three-dimensional protein structure through graph neural networks. Additionally, ReactEmbed \cite{sicherman2025reactembed} attempts to capture protein-molecule relationships through joint embedding training. While these methods provide rich protein representations, they all generate static embeddings that remain fixed regardless of the molecular context.

This comprehensive comparison allows us to assess the effectiveness of different protein representation strategies in enzymatic reaction prediction.

\subsection{Implementation Details}
\subsubsection{Problem Formulation}
Following previous works, the task is formulated as a sequence-to-sequence translation problem \cite{schwaller2019molecular,gricourt2024artificial,kreutter2021predicting}, where molecular structures are represented using SMILES (Simplified Molecular Input Line Entry System) notation. As shown in Figure \ref{fig:biocat}, the input molecule's SMILES string is ``translated'' into the product molecule's SMILES string using transformer-based architectures.

\subsubsection{Model Architecture}
For processing molecular structures, we employ the T5 architecture consisting of 6 encoder and decoder layers, with model dimension of 512, feed-forward dimension of 2048, and 8 attention heads. All SMILES sequences are tokenized and padded to a maximum length of 200 tokens. This chemical processing backbone remains consistent across all experimental configurations.

For methods incorporating protein information (ESM3, ProtBERT, GearNet, ReactEmbed, and our DAA), the protein representation is injected as a special token at the beginning of the encoder sequence, as illustrated in Figure \ref{fig:biocat}. 

\subsubsection{Training Protocol}
All models are trained on NVIDIA A40 GPUs using the AdamW optimizer with a learning rate of 1e-5 and batch size of 256. To ensure fair comparison, we maintain identical architectural configurations and training procedures across all experimental conditions, varying only the method of protein representation between our approach and the baselines. This standardization ensures that performance differences can be attributed specifically to the protein representation strategy rather than training dynamics or model capacity.

All code and pre-trained models are available at GitHub\footnote{https://anonymous.4open.science/r/DockingAwareAttention-8B8E} to ensure reproducibility.

%%%%%%%%%%%%%%%%%%%%%%%%%%%%%%%%%%%%%%%%%%%%%%%%%%%%%%%%%%%%%%%%
%%%%%%%%%%%%%%%%        MAIN RES  %%%%%%%%%%%%%%%%%%%%%%%%%%%
%%%%%%%%%%%%%%%%%%%%%%%%%%%%%%%%%%%%%%%%%%%%%%%%%%%%%%%%%%%%%%%%
\begin{table*}[!ht]\centering
\caption{Comprehensive performance comparison across different evaluation settings using top-k accuracy (\%). Results are shown for overall performance (All), high-complexity reactions (Complex), and innovative reactions (Novel). Higher percentages indicate better performance. Bold text indicates best performance with statistically significant improvement over the next best in each category.}
\label{tab:res-main}
\begin{tabular}{l|ccc|ccc|ccc}\toprule
& \multicolumn{3}{c|}{\textbf{All}} & \multicolumn{3}{c|}{\textbf{Complex}} & \multicolumn{3}{c}{\textbf{Novel}} \\
\textbf{Method} & \textbf{Top-1 (\%)} & \textbf{Top-3 (\%)} & \textbf{Top-5 (\%)} & \textbf{Top-1 (\%)} & \textbf{Top-3 (\%)} & \textbf{Top-5 (\%)} & \textbf{Top-1 (\%)} & \textbf{Top-3 (\%)} & \textbf{Top-5 (\%)} \\\midrule
Chemical-Only \cite{kreutter2021predicting} &37.29 &54.65 &61.39 &30.78 &44.92 &51.56 &18.25 &33.75 &40.82 \\
EC Tokens \cite{Probst2022} &46.01 &61.71 &66.64 &34.26 &51.22 &56.79 &31.42 &44.46 &49.45 \\
ESM3 \cite{hayes2024simulating} &44.58 &60.64 &65.62 &34.23 &48.57 &54.11 &27.39 &42.35 &47.47 \\
ProtBERT \cite{brandes2021proteinbert} &43.49 &59.52 &64.82 &33.11 &45.48 &50.42 &31.21 &40.12 &44.66 \\
GearNet \cite{zhang2023protein} &43.95 &60.10 &63.84 &28.09 &40.80 &46.23 &20.16 &32.62 &39.13 \\
ReactEmbed\cite{sicherman2025reactembed} &47.41 &61.19 &66.25 &36.90 &50.87 &55.18 &26.37 &42.68 &48.08 \\
DAA (Ours) &\textbf{49.96} &\textbf{66.65} &\textbf{71.48} &\textbf{41.62} &\textbf{56.43} &\textbf{62.20} &\textbf{35.24} &\textbf{50.66} &\textbf{55.54} \\
\bottomrule
\end{tabular}
\end{table*}
%%%%%%%%%%%%%%%%%%%%%%%%%%%%%%%%%%%%%%%%%%%%%%%%%%%%%%%%%%%%%%%%
%%%%%%%%%%%%%%%%      MAIN RES  %%%%%%%%%%%%%%%%%%%%%%%%%%%
%%%%%%%%%%%%%%%%%%%%%%%%%%%%%%%%%%%%%%%%%%%%%%%%%%%%%%%%%%%%%%%%

\subsection{Evaluation Protocol}
\label{sec:evaluation}
Following standard practice in reaction prediction literature \cite{Chen2023,kreutter2021predicting,schwaller2019molecular,gricourt2024artificial,Probst2022}, we evaluate models using Top-k accuracy metrics. In the context of chemical reaction prediction, where multiple valid products may exist for a given reaction, Top-k accuracy is particularly relevant as it captures the model's ability to propose reasonable reaction outcomes. Specifically, a prediction is considered correct if the true product appears among the model's k highest-confidence predictions.

\subsubsection{Statistical Significance Testing}
To assess the statistical significance of our results, we employ the Two-Proportion Z-Test. For this analysis, we consider each model prediction as a binary outcome (correct or incorrect), enabling direct comparison between different approaches. 
The Two-Proportion Z-Test evaluates whether the difference in success rates between two models is statistically significant by comparing their observed success rates and sample sizes.
For all reported performance improvements, we maintain a significance level of $\alpha = 0.05$.

\section{Empirical Results}
Our empirical results demonstrate the effectiveness of DAA across multiple evaluation scenarios. We analyze the performance on three key aspects: overall prediction accuracy, handling of complex molecules, and generalization to innovative reactions.
As discussed in Section \ref{sec:evaluation}, we evaluate performance using top-k accuracy.

Table \ref{tab:res-main} summarizes the results. DAA consistently outperforms all baseline approaches across all evaluation metrics. At K=1, DAA achieves 49.96\% accuracy, representing a significant improvement over both the Chemical-Only baseline (37.29\%) and methods using static protein representations such as ESM3 (44.58\%) and ProtBERT (43.49\%). Even compared to the strong ReactEmbed baseline (47.41\%), DAA shows a clear advantage with a 2.55 percentage point improvement.
The performance gap widens for higher K values, with DAA achieving 66.65\% at K=3 and 71.48\% at K=5, compared to ReactEmbed's 61.19\% and 66.25\% respectively. This consistent improvement across different K values suggests that DAA's dynamic protein representations provide valuable information for ranking potential reaction products.
Notably, the EC Tokens approach (46.01\% at K=1) performs better than sequence-based methods like ESM3 and ProtBERT, highlighting the value of enzyme classification information. However, DAA's superior performance (49.96\% at K=1) demonstrates that our context-aware representations capture more nuanced protein-substrate relationships than static EC number assignments.

\subsection{Performance on High-Complexity Reactions}
We evaluate performance on reactions involving complex molecules, defined using the Bertz Complexity Index \cite{bertz1981first}. For this analysis, we classify molecules with a Bertz Complexity Index exceeding 1500 as complex. These molecules present particular challenges for reaction prediction due to their numerous potential reaction sites and complex stereochemical considerations. When evaluating performance on these complex-molecule reactions (Table \ref{tab:res-main}), the advantages of DAA become even more pronounced. Our method achieves 41.62\% accuracy at K=1, substantially outperforming all baselines including ReactEmbed (36.90\%) and EC Tokens (34.26\%). This represents a nearly 13\% relative improvement over the next best method. The performance gap is particularly notable when compared to structure-based approaches like GearNet (28.09\% at K=1), suggesting that static structural representations alone are insufficient for handling complex molecular interactions. The significant drop in performance for all baselines on complex molecules (compared to the overall results) highlights the challenging nature of these cases, making DAA's robust performance especially valuable.

\subsection{Performance on Innovative Reactions}

Perhaps most significantly, DAA shows strong generalization capabilities when predicting innovative reactions - those involving target molecules not seen during training (Table \ref{tab:res-main}). Our method achieves 35.24\% accuracy at K=1 and 50.66\% at K=3, substantially outperforming both the Chemical-Only baseline (18.25\% and 33.75\%) and sophisticated protein representation methods like ESM3 (27.39\% and 42.35\%).
The performance gap between DAA and EC Tokens (31.42\% at K=1) is particularly noteworthy in this scenario, as it demonstrates that our dynamic representations better capture the underlying principles of enzyme-substrate interactions rather than simply memorizing known reaction patterns. This advantage in predicting novel transformations is crucial for practical applications in biocatalysis discovery and enzyme engineering.
Interestingly, structure-based methods like GearNet show relatively poor performance on innovative reactions (20.16\% at K=1), suggesting that static structural information alone may not generalize well to novel chemical transformations. The significant improvement achieved by DAA (35.24\% at K=1) validates our approach of combining physical interaction information through docking scores with learned attention patterns.

\subsection{Ablation Studies}
%%%%%%%%%%%%%%%%%%%%%%%%%%%%%%%%%%%%%%%%%%%%%%%%%%%%%%%%%%%%%%%%
%%%%%%%%%%%%%%%%        ABL RES  %%%%%%%%%%%%%%%%%%%%%%%%%%%
%%%%%%%%%%%%%%%%%%%%%%%%%%%%%%%%%%%%%%%%%%%%%%%%%%%%%%%%%%%%%%%%
\begin{table}[ht]
\caption{Ablation study results examining three key components: (1) Attention mechanism variants, comparing our full DAA approach against standard attention and docking-only alternatives, (2) Embedding model performance, comparing each model with and without DAA, and (3) Token integration strategies, evaluating different methods for incorporating protein representations into the sequence model. All results show top-k accuracy percentages.}
\label{tab:ablation-full}
\centering
\begin{tabular}{lccc}
\toprule
\textbf{Component} & \textbf{Top-1 (\%)} & \textbf{Top-3 (\%)} & \textbf{Top-5 (\%)} \\
\midrule
\multicolumn{4}{c}{Attention Mechanism} \\ \hline
Full DAA & \textbf{49.96} & \textbf{66.65} & \textbf{71.48} \\
Standard Attention & 46.55 & 61.41 & 67.47 \\
Docking-Only & 45.63 & 62.34 & 66.95 \\
\midrule
\multicolumn{4}{c}{Embedding Models} \\ \hline
\multicolumn{4}{l}{\textit{ESM3}} \\
\quad \quad Base Model  &44.58 &60.64 &65.62 \\
\quad \quad DAA Enhanced & \textbf{49.96} & \textbf{66.65} & \textbf{71.48} \\
\multicolumn{4}{l}{\textit{ProtBERT}} \\
\quad \quad Base Model &43.49 &59.52 &64.8 \\
\quad \quad DAA Enhanced &\textbf{ 45.72} & \textbf{61.68} & \textbf{67.64} \\
\multicolumn{4}{l}{\textit{GearNet}} \\
\quad \quad Base Model &43.95 &60.10 &63.84 \\
\quad \quad DAA Enhanced & \textbf{44.84} & \textbf{60.80} & \textbf{66.76} \\
\midrule
\multicolumn{4}{c}{Token Integration} \\ \hline
New Token & \textbf{49.96} & \textbf{66.65} & \textbf{71.48} \\
Concatenation & 49.51 & 62.69 & 66.77 \\
Addition & 40.23 & 57.113 & 63.49 \\
\bottomrule
\end{tabular}
\end{table}
%%%%%%%%%%%%%%%%%%%%%%%%%%%%%%%%%%%%%%%%%%%%%%%%%%%%%%%%%%%%%%%%
%%%%%%%%%%%%%%%%      ABL RES  %%%%%%%%%%%%%%%%%%%%%%%%%%%
%%%%%%%%%%%%%%%%%%%%%%%%%%%%%%%%%%%%%%%%%%%%%%%%%%%%%%%%%%%%%%%%

To thoroughly validate our approach and understand the contribution of each architectural component, we conducted three comprehensive ablation studies examining attention mechanisms, embedding model selection, and token integration strategies. Table~\ref{tab:ablation-full} presents the quantitative results across all variants, demonstrating the impact of each design choice on model performance.

\subsubsection{Attention Mechanism Analysis}
Our first ablation study evaluates different variants of the attention mechanism to understand the relative contribution of learned patterns versus physical interaction information. The results demonstrate the complementary nature of these information sources. While the standard attention mechanism achieves solid performance (67.47\% at top-5) by capturing sequence-based patterns, the docking-only approach (66.95\% at top-5) shows that physical interaction scores alone provide valuable signals for understanding protein-molecule relationships. The superior performance of our full DAA architecture (71.48\% at top-5) validates our hypothesis that combining both sources of information enables more effective modeling of protein-molecule interactions. This significant improvement over individual components suggests that learned attention patterns and physical docking information capture different aspects of protein behavior, leading to more comprehensive and accurate representations when integrated through our DAA mechanism.

\subsubsection{Impact of Embedding Models}
The second ablation study examines the influence of different protein embedding architectures on overall performance. Results show that incorporating DAA consistently improves performance across all embedding models tested. The base ESM3 model achieves 65.62\% top-5 accuracy as a baseline, improving substantially to 71.48\% with DAA enhancement. Similar improvements are observed with ProtBERT (64.8\% to 67.64\%) and GearNet (63.84\% to 66.76\%). These results demonstrate that DAA's benefits are not limited to a specific embedding architecture but rather represent a general improvement in protein representation learning. The consistent performance gains across different architectures (ranging from 2.9 to 5.9 percentage points) suggest that DAA's context-aware attention mechanism provides complementary information regardless of the base embedding model's architecture.

\subsubsection{Token Integration Strategies}
Our final ablation study investigates different methods for incorporating the protein representation vector into the sequence-to-sequence architecture. The substantial performance differences between integration strategies highlight the importance of proper protein information integration. Our new token approach achieves the best performance (71.48\% at top-5), significantly outperforming both concatenation (66.77\%) and addition (63.49\%) strategies. The poor performance of simple addition suggests that naive combination strategies can actually degrade model performance, while our new token approach allows the model to learn optimal integration patterns. The nearly 8 percentage point gap between the best and worst performing strategies emphasizes that the method of combining protein representations with molecular information significantly impacts the model's ability to leverage protein-specific context for reaction prediction.

\section{Discussion}

\subsection{Analysis of Attention Distribution Patterns}

%%%%%%%%%%%%%%%%        FIG DIST  %%%%%%%%%%%%%%%%%%%%%%%%%%%
%%%%%%%%%%%%%%%%%%%%%%%%%%%%%%%%%%%%%%%%%%%%%%%%%%%%%%%%%%%%%%%%
\begin{figure}[h]
    \includegraphics[width=0.45\textwidth]{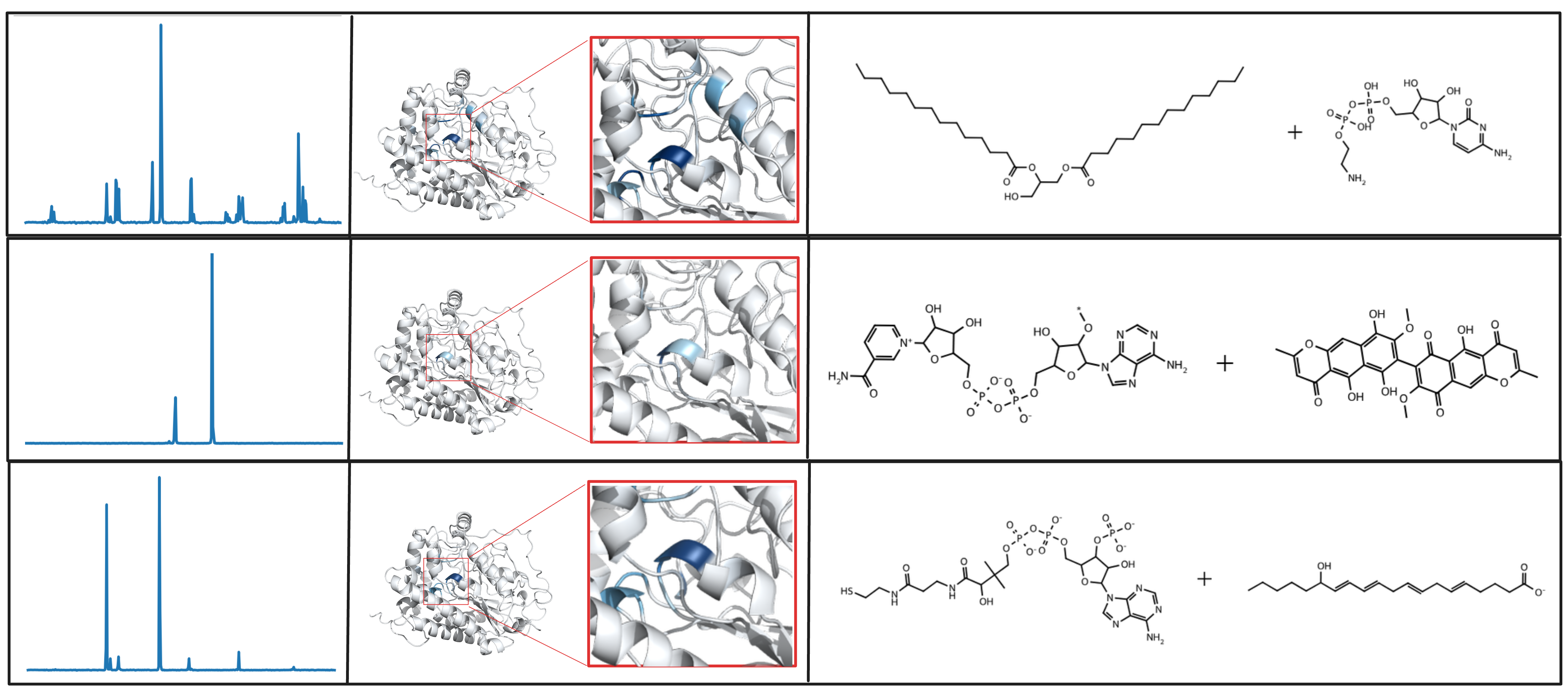}
    \caption{Attention patterns of triacylglycerol lipase (EC 3.1.1.3) across three reactions, showing: attention weights in sequence space (left), 3D structural visualization with attention intensity (center), and corresponding chemical reactions (right). The varying patterns demonstrate DAA's context-dependent adaptation.}
    \label{fig:exmp}
\end{figure}

%%%%%%%%%%%%%%%%%%%%%%%%%%%%%%%%%%%%%%%%%%%%%%%%%%%%%%%%%%%%%%%%
%%%%%%%%%%%%%%%%      FIG DIST  %%%%%%%%%%%%%%%%%%%%%%%%%%%
%%%%%%%%%%%%%%%%%%%%%%%%%%%%%%%%%%%%%%%%%%%%%%%%%%%%%%%%%%%%%%%%

Figure \ref{fig:exmp} illustrates the dynamic nature of DAA through a visualization of triacylglycerol lipase (EC 3.1.1.3) participating in three different reactions. The visualization reveals how our attention mechanism generates distinct patterns for the same enzyme when catalyzing different reactions, adapting its focus based on the specific molecular context. Each row in the figure presents a different reaction scenario, with the attention patterns visualized in both sequence space (1D attention profile) and structural space (3D protein representation), alongside the corresponding chemical transformation.

These visualizations provide compelling evidence that DAA successfully overcomes the limitations of static protein representations. By generating context-specific attention patterns, our method captures the inherent flexibility of enzyme behavior in different molecular environments. The apparent differences in attention distribution across different reactions, obvious in sequence profiles and structural visualizations, validate our core hypothesis about the importance of dynamic, interaction-aware protein representations in accurately modeling enzyme-catalyzed reactions.

%%%%%%%%%%%%%%%%%%%%%%%%%%%%%%%%%%%%%%%%%%%%%%%%%%%%%%%%%%%%%%%%

\subsection{Geometric Analysis of Protein Space}
%%%%%%%%%%%%%%%%%%%%%%%%%%%%%%%%%%%%%%%%%%%%%%%%%%%%%%%%%%%%%%%%
%%%%%%%%%%%%%%%%        FIG PCA  %%%%%%%%%%%%%%%%%%%%%%%%%%%
%%%%%%%%%%%%%%%%%%%%%%%%%%%%%%%%%%%%%%%%%%%%%%%%%%%%%%%%%%%%%%%%
\begin{figure}[h]
    \centering
    \includegraphics[width=1\linewidth]{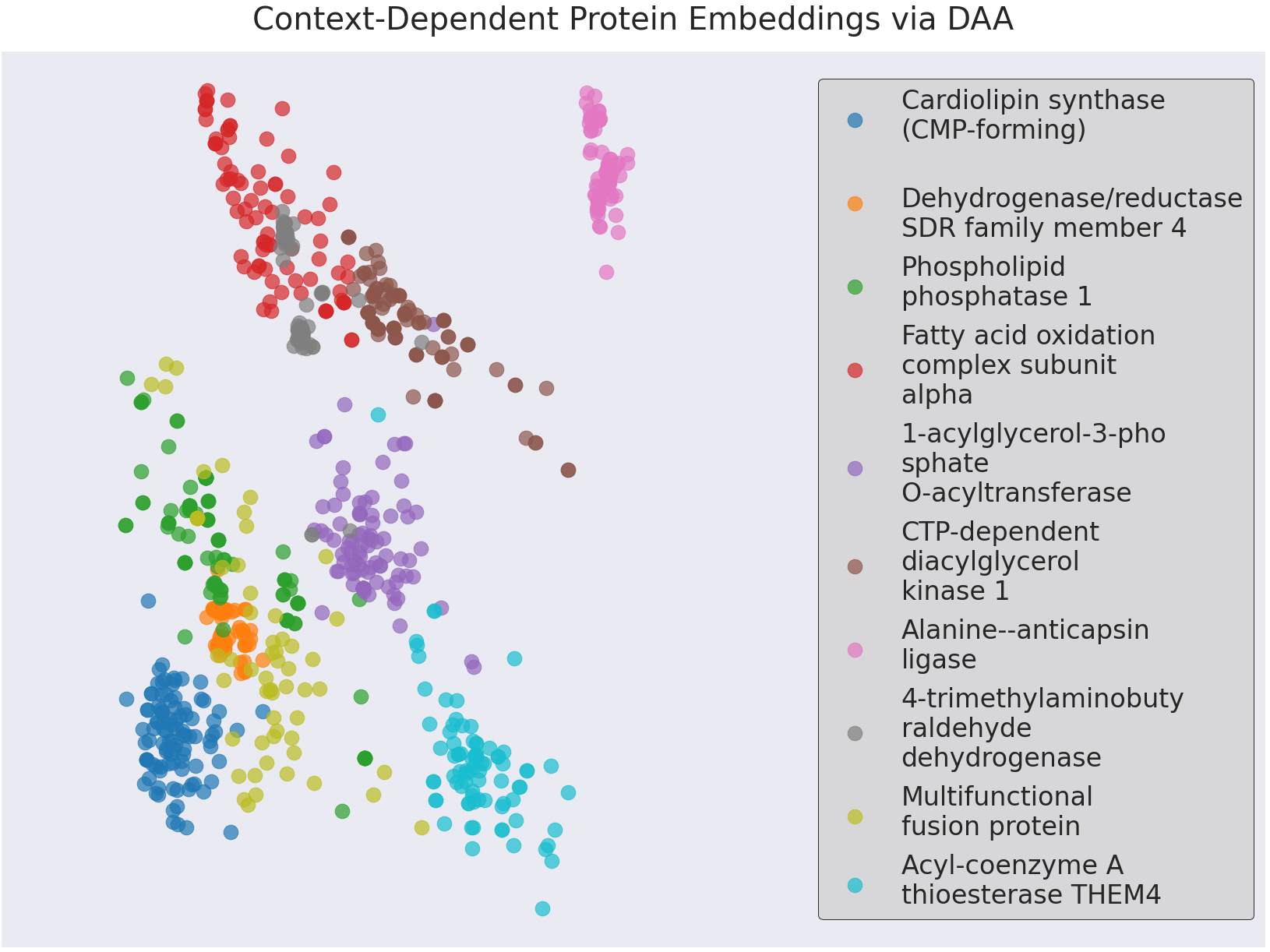}
    \caption{PCA visualization of DAA-generated protein embeddings, showing protein-specific clusters (colors) with intra-cluster variation. Each point represents a protein in a specific molecular context, demonstrating both preserved protein identity and context-dependent adaptation. Analysis covers 10 ECREACT proteins with ~50 molecular contexts each.}
    \label{fig:protein_embeddings}
\end{figure}

%%%%%%%%%%%%%%%%%%%%%%%%%%%%%%%%%%%%%%%%%%%%%%%%%%%%%%%%%%%%%%%%
%%%%%%%%%%%%%%%%      ABL PCA  %%%%%%%%%%%%%%%%%%%%%%%%%%%
%%%%%%%%%%%%%%%%%%%%%%%%%%%%%%%%%%%%%%%%%%%%%%%%%%%%%%%%%%%%%%%%

To understand the geometric properties of our learned protein representations, we analyzed the embedding space using Principal Component Analysis (PCA). We first trained PCA on all protein-substrate pairs in our dataset to learn the principal components. For visualization purposes, we then randomly selected 10 diverse proteins from ECREACT, each with approximately 50 different substrate interactions, and projected their embeddings onto the first two principal components.

Figure~\ref{fig:protein_embeddings} reveals several key properties of the learned representations: Despite their context-dependent nature, proteins maintain distinct identities by forming consistent clusters in the representation space, demonstrating DAA's ability to preserve protein-specific information while adapting to different molecular contexts. Within these clusters, we observe significant spread in the embedding positions, reflecting dynamic adaptations to different molecular contexts rather than collapsing to static, single-point representations. This intra-cluster variation illustrates how DAA generates flexible representations that respond to specific substrate interactions. Notably, while proteins generally maintain distinct clusters, we observe partial overlap between certain clusters in the embedding space, suggesting that DAA captures shared properties between different proteins when processing certain substrates. The biological basis for these inter-protein relationships merits further investigation.

These results demonstrate that DAA achieves a crucial balance: the embeddings maintain distinct protein identities while adapting to different molecular contexts. The organization of the embedding space shows both clear protein-specific clustering and meaningful variation based on molecular interactions, supporting DAA's effectiveness in generating dynamic protein representations.

\section{Conclusion}
This work introduces Docking-Aware Attention (DAA), a novel architecture that addresses a fundamental challenge in enzymatic reaction prediction: capturing the context-dependent nature of enzyme-substrate interactions. By incorporating molecular docking information into the attention mechanism, DAA generates dynamic protein representations that adapt based on specific molecular interactions. Our comprehensive evaluation demonstrates significant improvements over existing methods, particularly in challenging scenarios involving complex molecules and innovative reactions.
 
 The success of DAA in biocatalysis prediction validates our core hypothesis that protein representations should adapt to different molecular contexts. The substantial performance improvements—achieving 62.2\% accuracy versus 56.79\% baseline for complex molecules and 55.44\% versus 49.45\% for innovative reactions demonstrate the practical value of incorporating physical interaction information into reaction prediction systems. Our ablation studies further confirm that each architectural component contributes meaningfully to the final performance, with the combination of learned attention patterns and docking-based interaction scores proving particularly effective.

While our results are promising, several directions for future research emerge:  (i) \textit{Synthetic Route Planning:} Integrating DAA into retrosynthesis planning systems could enable more accurate prediction of feasible enzymatic transformations in multi-step synthesis pathways. This could lead to more efficient routes for complex molecule synthesis by better accounting for enzyme-substrate compatibility at each step; (2) \textit{Reaction Condition Optimization:} The DAA framework could be extended to predict optimal reaction conditions by modeling how enzyme-substrate interactions change under different temperature, pH, and solvent conditions. This could improve reaction yield and selectivity in practical applications;

By open-sourcing our implementation and pre-trained models, we aim to facilitate further research in computational biocatalysis and encourage the development of increasingly sophisticated approaches to enzymatic reaction prediction. DAA's success demonstrates the value of combining physical insights with modern deep learning architectures, pointing toward a promising direction for improving predictive models in chemical synthesis.

\bibliographystyle{ACM-Reference-Format}
\bibliography{ref}

\end{document}